\pdfoutput=1

\documentclass[11pt]{article}
\usepackage{booktabs} 
\usepackage{hyperref}
\usepackage{csquotes}
\usepackage{float}
\usepackage{acl}

\usepackage{times}
\usepackage{latexsym}
\usepackage{enumitem}
\usepackage[T1]{fontenc}

\usepackage[utf8]{inputenc}

\usepackage{microtype}

\title{Beyond Black Box AI-Generated Plagiarism Detection: From Sentence to Document Level}

\usepackage{graphicx}
\usepackage{url}
\usepackage{booktabs}
\usepackage{amssymb}

\author{Mujahid Ali Quidwai \\
  New York University \\
  \texttt{maq4265@nyu.edu} \\
\And
Chunhui Li \\
  Columbia University \\
  \texttt{cl4282@columbia.edu} \\
\And
Parijat Dube \\
  IBM Research \\
  \texttt{pdube@us.ibm.com} \\}

\begin{document}
\maketitle
\begin{abstract}
The increasing reliance on large language models (LLMs) in academic writing has led to a rise in plagiarism. Existing AI-generated text classifiers have limited accuracy and often produce false positives. We propose a novel approach using natural language processing (NLP) techniques, offering quantifiable metrics at both sentence and document levels for easier interpretation by human evaluators. Our method employs a multi-faceted approach, generating multiple paraphrased versions of a given question and inputting them into the LLM to generate answers. By using a contrastive loss function based on cosine similarity, we match generated sentences with those from the student's response. Our approach achieves up to 94\% accuracy in classifying human and AI text, providing a robust and adaptable solution for plagiarism detection in academic settings. This method improves with LLM advancements, reducing the need for new model training or reconfiguration, and offers a more transparent way of evaluating and detecting AI-generated text.
\end{abstract}

\section{Introduction}
In recent years, large language models (LLMs) have demonstrated remarkable capabilities across a wide range of natural language processing (NLP) tasks, including text classification, sentiment analysis, translation, and question-answering \citep{MGTBench}.

These foundational models exhibit immense potential in tackling a diverse array of NLP tasks, spanning from natural language understanding (NLU) to natural language generation (NLG), and even laying the groundwork for Artificial General Intelligence (AGI) \citep{harnessing}. In the world of advanced LLMs, \citet{chatgpt} as an AI model developed by \citet{openai} has become one of the most popular and widely used models, setting new records for performance and flexibility in many applications. According to the latest available data, \citet{chatgpt} currently has over 100 million users and the website currently generates 1 billion visitors per month \citep{users}. While ChatGPT has brought numerous benefits such as it allows us to obtain information more effectively, improves people's writing skills etc., however, it has also introduced considerable risks \cite{openai2023gpt4}.

A major risk associated with the growing dependence on ChatGPT is the escalation of plagiarism in academic writing \cite{khalil2023chatgpt}, which subsequently compromises the integrity and purpose of assignments and examinations. Thanks to its advanced training process and access to abundant pre-training data sets, ChatGPT is capable of resembling human-like language when provided with a prompt \citep{blessing2023}. It even exceeds human performance in some academic writing while maintaining authenticity and richness. Furthermore, humans are unable to accurately distinguish between Human Generated Text (HGT) and Machine Generated Text (MGT), regardless of their familiarity with ChatGPT \citep{comparison}. These factors present significant challenges in maintaining educational integrity and challenge the current paradigm of how teachers teach.
\begin{figure}[h]
    \includegraphics[width=0.48\textwidth]{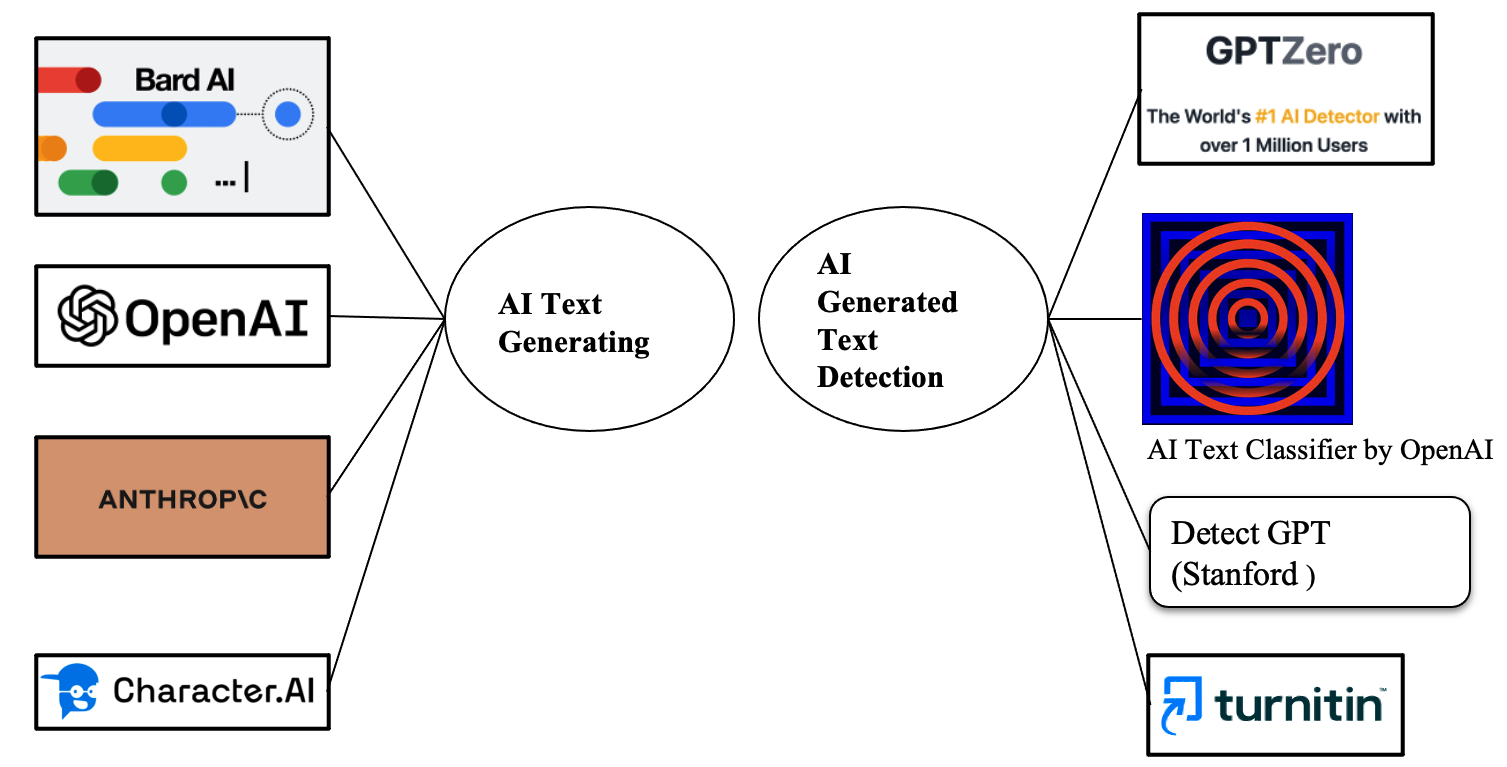}

    \caption{Popular LLMs and AI-generated text detection tools}
    \label{figure 1}
\end{figure}

To reduce potential plagiarism caused by the use of LLMs, researchers have developed various AI-generated text classifiers or tools such as  
Log-Likelihood~\cite{loglikelihood},
RoBERTa-QA (HC3)~\cite{guo2023close},~\citet{gptzero},
OpenAI Classifier~\cite{openai},
DetectGPT~\cite{mitchell2023detectgpt},
and Turntin~\cite{innocent}. Figure~\ref{figure 1} lists popular LLMs used for text generation and AI-generated text detection tools.

Existing approaches for detecting text generated by language models have several limitations as highlighted in Table~\ref{table 1}.
\begin{table}[h]
\centering
\begin{tabular}{lccccc}
\toprule
\textbf{Current Method} & \rotatebox{90}{\textbf{High False Positive}} & \rotatebox{90}{\textbf{Model Retraining}} & \rotatebox{90}{\textbf{Works for GPT2}} & \rotatebox{90}{\textbf{Blackbox Nature}} \\
\midrule
\verb|Log-Likelihood| & $\checkmark$ & $\checkmark$ & $\checkmark$ & $\checkmark$ \\
\verb|RoBERTa-QA (HC3)| & $\checkmark$ & $\checkmark$ & $\times$ & $\checkmark$ \\
\verb|OpenAI Classifier| & $\checkmark$ & $\checkmark$ & $\times$ & $\checkmark$  \\ 
\verb|DetectGPT| & $\checkmark$ & $\checkmark$ & $\checkmark$ & $\checkmark$ \\ 
\verb|Turnitin| & $\checkmark$ & $\checkmark$ & $\times$&$\checkmark$ \\
\bottomrule
\end{tabular}
\caption{Problems with current approaches}
\label{tab:rotated_headers}
\label{table 1}
\end{table}
For instance, these tools may rapidly become outdated due to technological advancements, such as new versions of GPT models, necessitating classifier retraining and often resulting in limited accuracy. Models trained specifically on one language model might not effectively detect text generated by a different language model (e.g., DetectGPT classifier works only on text generated using GPT2 \cite{openai}). Additionally, some detection tools provide non-quantitative label results, and all possess a black-box nature concerning prediction accuracy. Thus the predictions made by such tools lack explainability and are challenging for human evaluators to comprehend. This issue leads to a high number of false positive punishments in academic settings \cite{innocent}.

We propose a novel approach for detecting plagiarized text, which focuses on NLP techniques. Our approach offers more quantifiable metrics at the sentence level, allowing for easier interpretation by human evaluators and eliminating the black-box nature of existing AI text detection methods. Our approach is not limited to ChatGPT but can also be applied to other LLMs such as BardAI \cite{bardai}, Character.AI \cite{character.ai}, and so on, it also can adapt automatically as those LLMs upgrade. This adaptability helps to ensure that it does not become outdated quickly as technology advances.

 In evaluating our approach, we used the open dataset known as the ChatGPT Comparison Corpus (HC3) \cite{guo2023close}. This dataset contains 10,000 questions and their corresponding answers from both human experts and ChatGPT, covering a range of domains including open-domain, computer science, finance, medicine, law, and psychology. Our approach achieves  \textbf{94\%} accuracy in classifying between human answers and ChatGPT answers in the HC3 data set.
 

The paper is structured as follows. Section 2 contains a review of relevant literature. Our proposed end-to-end approach for AI-text detection is detailed in Section 3, where we describe the method framework. In Section 4, we present our main results from the experimental evaluation. Lastly, we summarize our findings and discuss future directions in Section 5, which serves as the conclusion.

\section{Related Work}
The field of AI-generated text detection has garnered significant interest, but only a few models and tools have achieved widespread adoption. In this section, we discuss state-of-the-art approaches, the datasets used for training their classifiers, and their limitations.

\subsection{DetectGPT}
DetectGPT \cite{mitchell2023detectgpt} is a zero-shot machine-generated text detection method that leverages the negative curvature regions of an LLM's log probability function. The approach does not require training a separate classifier, collecting a dataset of real or generated passages, or watermarking generated text. Despite its effectiveness, DetectGPT is limited to GPT-2 generated text, and its performance may not extend to other LLMs~\citep{tang2023science}.

\subsection{Human ChatGPT Comparison Corpus (HC3)}
\citet{guo2023close} introduced the HC3 dataset, which contains tens of thousands of comparison responses from both human experts and \citet{chatgpt}. They conducted comprehensive human evaluations and linguistic analyses to study the characteristics of ChatGPT's responses, the differences and gaps from human experts, and future directions for LLMs. Furthermore, they built three different detection systems to effectively detect whether a text is generated by ChatGPT or humans. However, this approach might still suffer
from high false positive rates and it does not provide correct sentence-level comparison metrics.

\subsection{OpenAI AI Text Classifier}
The OpenAI AI Text Classifier \cite{openai} is a fine-tuned GPT model designed to predict the likelihood of a piece of text being AI-generated. This free tool aims to foster discussions on AI literacy, but it has limitations: it requires a minimum of 1,000 characters, can mislabel AI-generated and human-written text, and can be evaded by editing AI-generated text. Additionally, it also suffers from high positive rates.

In our research, we aim to address the limitations of these existing methods by developing a novel approach for detecting plagiarized text, focusing on natural language processing techniques that provide more quantifiable metrics and eliminate the black-box nature of existing AI text detection methods
\section{Our Method}
\label{sec:method}

\label{sec:method}
\begin{figure*}[ht]
    \centering
    \includegraphics[width=\textwidth]{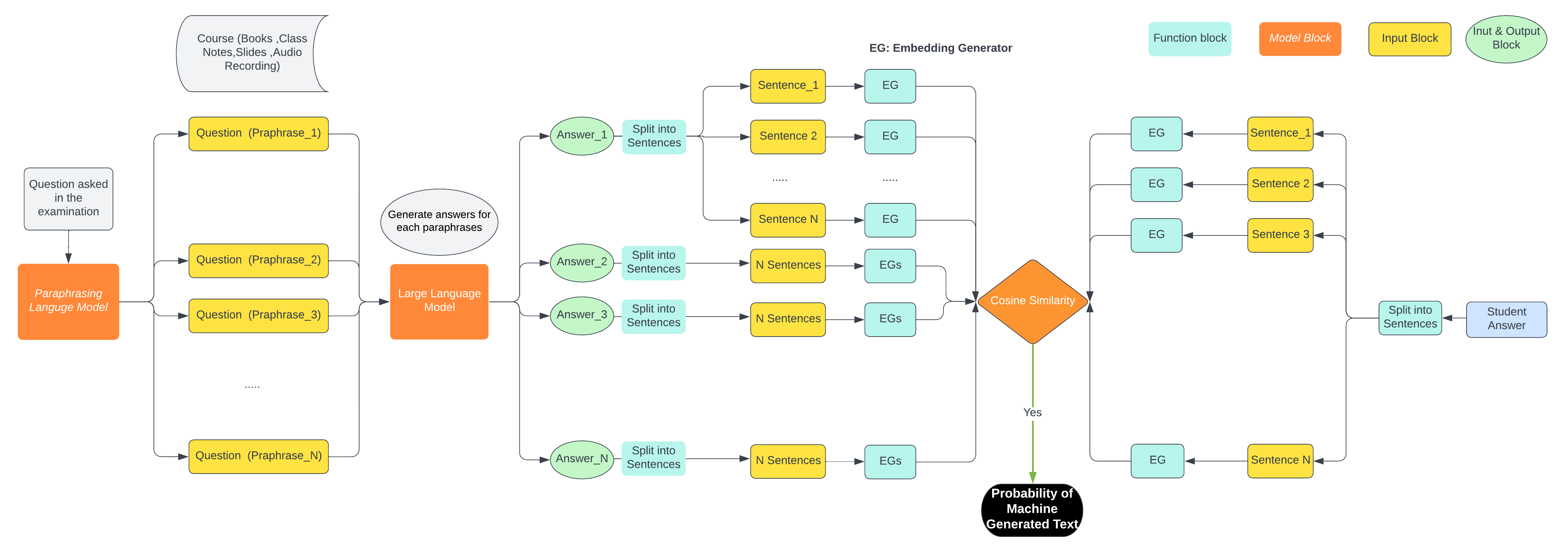}
    \caption{Model Architecture for our proposed method}
    \label{figure 2}
\end{figure*}

In this section, we present our approach to effectively compare and detect plagiarism in student responses. Our method utilizes an advanced paraphrasing model, a state-of-the-art language model, and a contrastive loss function to deliver a comprehensive and transparent evaluation system. Figure~\ref{figure 2} shows the different components of our proposed model architecture.

\subsection{Paraphrasing Model}

To simulate the variety of questions a student might pose to a large language model (LLM), we employ a paraphrasing model (refer to Figure~\ref{figure 2}). This model generates multiple paraphrased versions of a given question, accounting for the diversity in student queries and ensuring robustness in the detection process. 

 Using Google's T5 language model \cite{t5blog2020} an initial dataset of questions is selected and paraphrased. The T5 model is trained on a large corpus of text, which enables it to understand the context and rephrase questions that preserve the original meaning while introducing variety. This model mimics the array of questions students might ask an LLM, a crucial feature to ensure robustness in the detection process.

For example, consider the following original question Q1 from the Reddit ELI5 (HC3 dataset) \cite{guo2023close}:

\renewcommand{\labelenumi}{Q\theenumi}
\hypertarget{Q1}{}
\begin{enumerate}
    \item \label{Q1}What is a hacker really doing? I've always wanted to know why hackers are jamming on their keyboards and what they can possibly be doing that requires such precision and speed. Please explain like I'm five.
\end{enumerate}
The paraphrasing model generates the following paraphrased versions:

\renewcommand{\labelenumi}{P\theenumi}
\begin{enumerate}
    \item \label{P1} What do hackers actually do? I've always been curious about why they type so fast on their keyboards and what they're doing that needs such accuracy and quickness. Can you explain it to me as if I were a five-year-old?
    
    \item \label{P2} I've always wondered what hackers are truly up to when they're typing rapidly on their keyboards. What kind of tasks are they performing that demand such skill and swiftness? Please explain it in a simple way, as if I were just five years old.
    
    \item \label{P3} What is it that hackers do when they're furiously typing on their keyboards? I'm curious about the activities they're involved in that require such speed and precision. Could you break it down for me in terms a five-year-old would understand?
\end{enumerate}

\subsection{LLM Integration}
Once we have the paraphrased questions, we input them into an LLM. This model, particularly the GPT-3.5-turbo from OpenAI~\citet{chatgpt}, is adept at generating coherent and contextually appropriate answers. Its proficiency stems from pre-training on an extensive amount of textual data, allowing it to provide accurate and relevant responses to the paraphrased questions.

For example, we input the first paraphrased question \hyperlink{P1}{P1} into the LLM and generate the following answer:

\renewcommand{\labelenumi}{A\theenumi}
\hypertarget{A1}{}
\begin{enumerate}
    \item \label{A1} Hackers are like computer experts who solve puzzles. They use their keyboards to give commands to computers to find secret information or fix problems. They need to be fast and accurate because computers follow instructions very quickly, and one wrong command can cause mistakes. Just like playing a game, they need to be good at using their keyboards to win the computer puzzle.

\end{enumerate}

We do similar generations for the other two paraphrased versions of the question~\hyperlink{Q1}{Q1}.

\subsection{Evaluation Process}

To facilitate a detailed comparison between the LLM-generated answers and student responses, we break down each answer into individual sentences. This granular approach enhances transparency and allows for a more in-depth evaluation of potential plagiarism. 


For example, consider the  LLM-generated answer \hyperlink{A1}{A1} and a human answer~\hyperlink{H1}{H1} for question~\hyperlink{Q1}{Q1} from the Reddit ELI5 dataset:
\renewcommand{\labelenumi}{H\theenumi}
\hypertarget{H1}{}
\begin{enumerate}
    \item \label{H1} I've always wanted to know why hackers are jamming on their keyboards In reality, this doesn't happen. This is done in movies to make it look dramatic and exciting. Real computer hacking involves staring at a computer screen for hours of a time, searching a lot on Google, muttering \" hmmm \" and various expletives to oneself now and then, and stroking one 's hacker - beard while occasionally tapping on a few keys .", "Computers are stupid, they don't know what they are doing, they just do it. If you tell a computer to give a cake to every person that walks through the door, it will do. Hackers are the people that get extra cake by going around the building and back through the door. GLaDOS however, will give you no cake .", "Hackers have a deep and complete understanding of a subject ( e.g. a machine or computer program ). They change the behavior of the subject to something that was never intended or even thought it would be possible by the creator of the subject .

\end{enumerate}

We next do a pair-wise comparison between a sentence in \hyperlink{H1}{H1} and all the sentences in \hyperlink{A1}{A1}, {A2}, and {A3}, to identify the AI generated sentence which is most similar to  ~\hyperlink{H1}{H1}.

\subsection{Cosine Similarity}
To compare two sentences we measure cosine similarity between the embeddings for the sentences generated using  \texttt{text-embedding-ada-002}\citet{}. The use of cosine similarity on sentence level contextualembeddings captures semantic and syntactic congruence between compared sentences. We use the term Human-Machine (HM) comparison for comparing sentence pairs involving a human-generated sentence and a machine-generated sentence. While Machine-Machine (MM) comparison involves comparing two machine-generated sentences.




\subsection{Linear Discriminant Analysis}

We apply Linear Discriminant Analysis (LDA) \cite{tharwat2017lda} —a supervised classification method — to categorize sentences as human- or AI-generated using cosine similarity scores. These scores and their respective category labels form our dataset, serving as independent and dependent variables, respectively.
The LDA model is trained using sklearn's \texttt{LinearDiscriminantAnalysis} class. The trained model is then used to predict the probability of a sentence in the test set being AI-generated.

To optimize classification, we explore a range of threshold values from 0 to 1 in a binary system. By assigning samples in datasets HM and MM to categories 0 and 1 respectively, we can conduct the LDA analysis on these two groups of datasets. Consequently, we determine the optimal threshold for classifying human-generated text and AI-generated text awhere the accuracy is maximized.
\label{sec:experiment}
\label{sec:experiment}
\section{Experimental Evaluation}
In our experimental evaluation, we aim to measure the accuracy of our approach in detecting similarities between human and machine-generated answers. We use the Human ChatGPT Comparison Corpus (HC3) dataset, which contains human and ChatGPT-generated answers to the same questions.
\subsection{Dataset Preparation}
For our analysis, we prepare two datasets to evaluate our model at the sentence and document levels. We use the HC3 dataset for sentence-level evaluation and then we did a summation over sentence-level cosine similarity to get the average similarity for the document. Further, to evaluate generalization performance of our model, we use GPT-wiki-intro dataset~\citep{aaditya_bhat_2023} for document-level evaluation and comparison with other models.

\subsubsection{Sentence-level Dataset: HC3}
We first use the HC3 dataset, which contains questions and corresponding human and machine responses. The HC3 dataset has an additional machine response for each question, resulting in two machine-generated answers. 

Next, we break down each answer for a given question into individual sentences, creating a dataset of roughly 43,000 sentence-level comparisons for machine-machine (MM) and human-machine (HM) categories. 
We use this dataset to compare the human response to the machine response at the sentence level, as well as compare the machine responses to each other at the sentence level using cosine similarity. Some example cosine similarity values for HM and MM categories are presented in Table~\ref{tab:combined}. 
\begin{table}[ht]
\centering
\begin{tabular}{|c|c|c|c|}
\hline
\multicolumn{2}{|c|}{HM} & \multicolumn{2}{c|}{MM} \\
\hline
CS & Label & CS & Label \\
\hline
0.785 & 0 & 0.846 & 1 \\
0.826 & 0 & 0.824 & 1 \\
0.690 & 0 & 0.827 & 1 \\
0.778 & 0 & 0.824 & 1 \\
0.899 & 0 & 0.824 & 1 \\
\hline
\end{tabular}
\caption{Example results of cosine similarity (CS) on HM and MM sample with their corresponding categorical label (0,1)}
\label{tab:combined}
\end{table}



\begin{figure}[h]
    \includegraphics[width=0.45\textwidth]{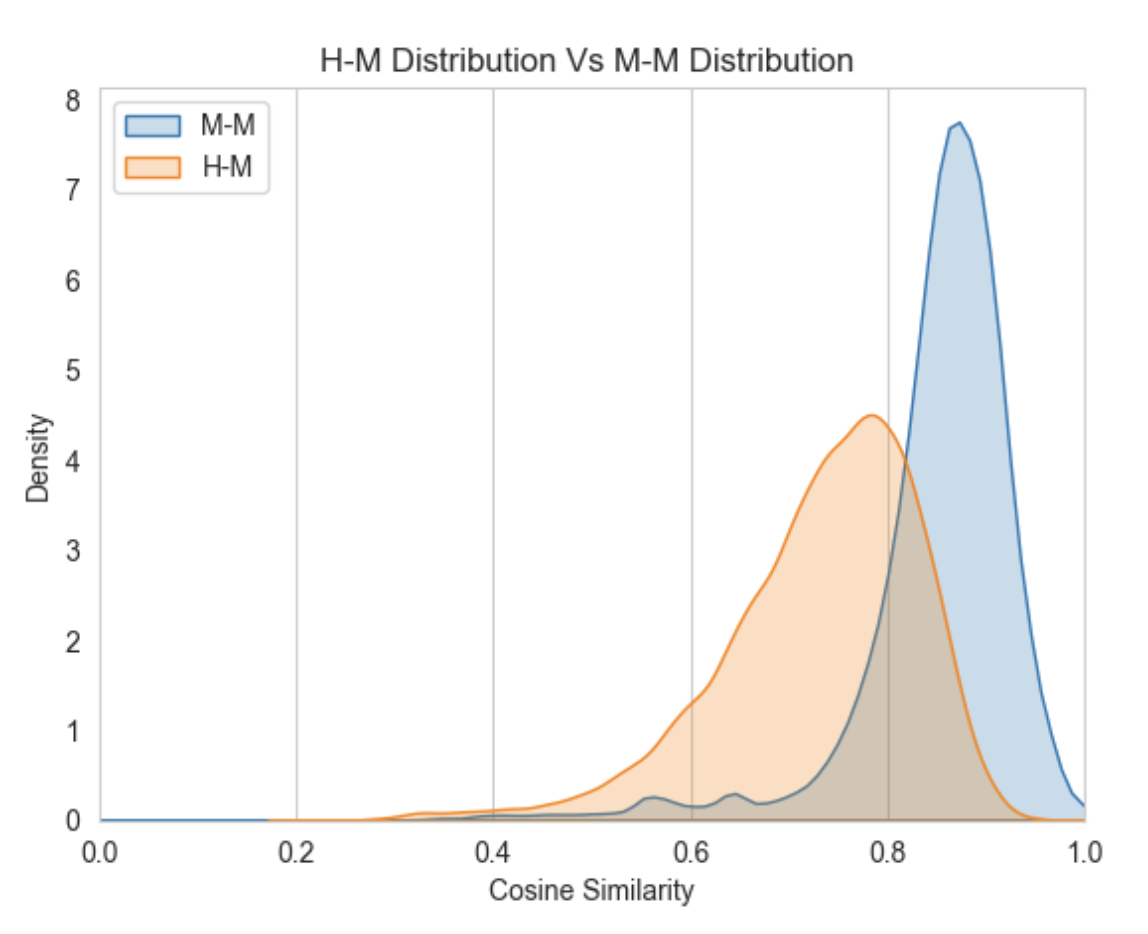}

    \caption{Distribution of cosine similarity at sentence level for HM and MM.}
    \label{figure: sentence level dist}
\end{figure}

Figure~\ref{figure: sentence level dist} shows the distribution of cosine similarity for HM and MM. For a visual representation of the cosine similarity scores distribution, we generate a Kernel Density Estimation (KDE) plot~\cite{chen2017tutorial}. We also calculate the mean and standard deviation of these scores (see Table~\ref{tab:cosine-similarity-statistics-sentence}) for sentence level in HM and MM samples, providing insights into the data. While the mean of the two classes is significantly different, they also have high standard deviations. This dataset is to be used to train and test our LDA model at the sentence level. 

\begin{table*}[h!]
\centering
\begin{tabular}{|c|c|c|}
\hline
\textbf{Statistic} & \textbf{Human-Machine (HM)} & \textbf{Machine-Machine (MM)} \\
\hline
Mean & 0.7309 & 0.8527 \\
\hline
Standard Deviation & 0.1016 & 0.0813 \\
\hline
\end{tabular}
\caption{Sentence Level Cosine Similarity Statistics}
\label{tab:cosine-similarity-statistics-sentence}
\end{table*}

Table~\ref{tab:sentence-level} shows the threshold value used in the LDA classifier and the corresponding accuracy on the test set.

\begin{table}
  \centering
  \begin{tabular}{|c|c|}
    \hline
    \textbf{LDA Model Result} & \textbf{Value} \\
    \hline
    Best Threshold & 0.40 \\
    \hline
    Accuracy & 0.80 \\
    \hline
  \end{tabular}
  \captionof{table}{LDA Model Results: Sentence level}
  \label{tab:sentence-level}
\end{table}

\subsubsection{Document-Level Dataset: HC3}
For a comprehensive understanding, we also conduct a document-level analysis utilizing the HC3 dataset. Rather than dissecting the responses into separate sentences, this level of examination treats the entire response as a single unit.

The document-level dataset is constructed by averaging the highest cosine similarity scores from the sentence-level comparison within each response. This approach ensures that the most closely matched sentences significantly impact the document-level similarity metric, thereby emphasizing the presence of highly similar sentences in the text. This similarity value serves as the foundation for our LDA model at the document level, allowing for a macroscopic comparison of the machine and human responses. 

\begin{figure}[h]
    \includegraphics[width=0.5\textwidth]{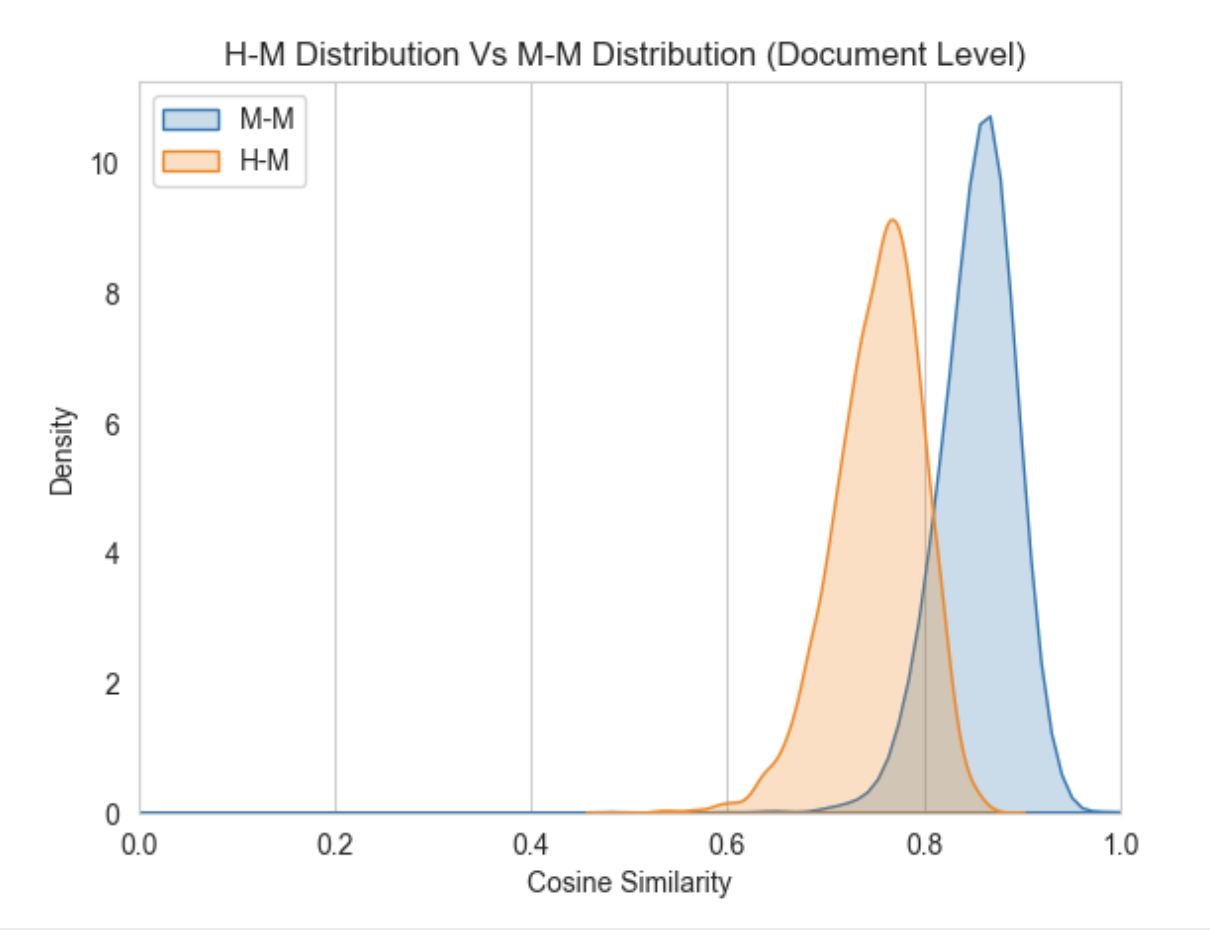}
    \caption{Distribution of cosine similarity at the Document level for HM and MM}
    \label{figure: document level lda}
\end{figure}

The distribution of cosine similarity at the document level is shown in Figure~\ref{figure: document level lda}. Table~\ref{tab:cosine-similarity-statistics-document} provides the mean and standard deviation of cosine similarity scores for HM and MM samples in the document level dataset.

\begin{table*}[h!]
\centering
\begin{tabular}{|c|c|c|}
\hline
\textbf{Statistic} & \textbf{Human-Machine (HM)} & \textbf{Machine-Machine (MM)} \\
\hline
Mean & 0.7343 & 0.8527 \\
\hline
Standard Deviation & 0.0447 & 0.0681 \\
\hline
\end{tabular}
\caption{Document Level Cosine Similarity Statistics}
\label{tab:cosine-similarity-statistics-document}
\end{table*}

The LDA classifier's threshold value and the corresponding accuracy on the test set for the document-level analysis are presented in Table~\ref{tab:document-level}. Observe that, in contrast to sentence level statistics (Table~\ref{tab:cosine-similarity-statistics-sentence}), the standard deviation of the two classes under document level comparison (Table~\ref{tab:cosine-similarity-statistics-document}) are smaller thereby resulting in a more discriminant classifier.

\begin{table}
  \centering
  \begin{tabular}{|c|c|}
    \hline
    \textbf{LDA Model Result} & \textbf{Value} \\
    \hline
    Best Threshold & 0.66 \\
    \hline
    Accuracy & 0.94 \\
    \hline
  \end{tabular}
  \captionof{table}{LDA Model Results: Document level}
  \label{tab:document-level}
\end{table}

\subsection{Experimental Setup}
Using the prepared dataset, we conduct a series of experiments to assess the performance of our proposed method in various plagiarism scenarios.

All the elements from the test set i.e., questions and corresponding student answers, including original and paraphrased questions alongside their corresponding AI-generated answers, are subsequently stored in a vector database, more specifically, Milvus~\citep{2021milvus}, an open-source vector database. This step ensures efficient data management, comparison, and high-speed searching of vector data.
We incorporate FastText, a module developed by Facebook~\citep{bojanowski2017enriching}, for vector ranking. The vectors representing paraphrased answers are ranked, creating a hierarchy of sentences based on similarity. A vector embedding generator from OpenAI aids in transforming the text into numerical form, allowing machine learning algorithms to process it. This transformation is pivotal for comparing student responses with AI-generated answers.

\subsection{Results and Analysis}
From Table ~\ref{tab:sentence-level} and Table ~\ref{tab:document-level} we observe that the LDA classifier works better at the document level compared to the sentence level.
We next conduct the document level evaluation of our model on the GPT-wiki-intro dataset. This dataset comprises questions along with their corresponding GPT-2 generated introductions and human-written introductions from Wikipedia articles. We perform document-level analysis on the first 100 examples from the GPT-wiki-intro dataset by comparing the AI-generated introductions to the human-written introductions, as well as comparing the AI-generated introductions to each other.

Our evaluation aims to demonstrate the explainability of our tool and its ability to provide both sentence and document level analysis. By comparing our results with existing benchmarks, we highlight the advantages of our approach in detecting plagiarism more effectively and transparently.


Our model is compared with two state-of-the-art (SOTA) models: HC3 and OpenAI's text classifier.
In order to evaluate the effectiveness of using the proposed paraphrasing model, we used two versions of our model, a model without paraphrasing (\textbf{A}) and a model employing paraphrasing (\textbf{B}) on the test set.
This allows us to directly assess the impact of paraphrasing on model performance.

Confusion matrices for all the models under evaluation are shown in Table~\ref{table:confusion_matrices}. While derived performance metrics (F1 score, precision, and recall) are provided in  Table~\ref{table:results}. 
We observe no improvement in model performance with paraphrasing  on this data set. We plan to investigate other potential approaches to improve model performance including varying the temperature and P value~\cite{openai2023gpt4} of LLM used for answer generation. We also plan to study our model performance on other datasets for a robust evaluation of the value of paraphrasing.


\begin{table}[h!]
\centering
\begin{tabular}{@{}lcc@{}}
\toprule
\textbf{} & \textbf{Predicted 0} & \textbf{Predicted 1} \\ \midrule
\multicolumn{3}{c}{RoBERTa-QA} \\ \midrule
\textbf{Actual 0} & 91 & 9 \\
\textbf{Actual 1} & 77 & 23 \\ \midrule
\multicolumn{3}{c}{OpenAI Classifier} \\ \midrule
\textbf{Actual 0} & 64 & 36 \\
\textbf{Actual 1} & 98 & 2 \\ \midrule
\multicolumn{3}{c}{\textbf{Our Model-A}} \\ \midrule
\textbf{Actual 0} & 99 & 1 \\
\textbf{Actual 1} & 90 & 10 \\ \midrule
\multicolumn{3}{c}{\textbf{Our Model-B}} \\ \midrule
\textbf{Actual 0} & 98 & 2 \\
\textbf{Actual 1} & 91 & 9 \\
\bottomrule
\end{tabular}
\caption{Confusion matrices for SOTA models and our model tested on GPT-wiki-intro dataset. Our model performance on Class 0 is better than both RoBERT-QA and Open AI Classifier, while on Class 1 our performance is better than RoBERTa-QA. Our Model-B uses paraphrasing.}
\label{table:confusion_matrices}
\end{table}

\begin{table}[h!]
\centering
\begin{tabular}{@{}lccc@{}}
\toprule
\textbf{Model} & \textbf{Precision} & \textbf{Recall} & \textbf{F1} \\ \midrule
RoBERTa-QA & 0.91               & 0.54            & 0.68            \\
OpenAI Classifier   & 0.64               & 0.39            & 0.49             \\
\textbf{Our Model-A}        & 0.99               & 0.52            & 0.69   
        \\
\textbf{Our Model-B}        & 0.98              & 0.52            & 0.68 
\\ \bottomrule
\end{tabular}
\caption{Document level F1 score, precision, and recall of the models. Our Model-B uses paraphrasing.}
\label{table:results}
\end{table}


Our model provides the probability of a text being AI-generated, both at sentence and document levels, enhancing transparency for evaluators examining potential plagiarism. For each sentence in a test document - in this case, a student response - the model calculates the probability of that sentence being LLM-generated. When utilizing the Reddit ELI5 (HC3 dataset), our model contrasts the human response with the LLM response on a sentence-by-sentence basis, as demonstrated in Table~\ref{tab:appendix-table}.
\begin{table*}[ht]
\centering
\begin{tabular}{|p{0.3\textwidth}|p{0.3\textwidth}|c|}
\hline
\textbf{LLM Response} & \textbf{Human Response} & \textbf{Cosine Similarity} \\
\hline
This can involve a lot of trial and error, which is why hackers might seem to be "jamming on their keyboards" as they try different approaches. & Computers are stupid , they do n't know what they are doing , they just do it. & 0.8087 \\
\hline
Overall, hacking can be a complex and technical activity that requires a lot of knowledge and skill. & Hackers have a deep and complete understanding of a subject (e.g., a machine or computer program). & 0.8753 \\
\hline
Hacking can involve a lot of typing and computer use, because hackers often use special software and programs to try to find weaknesses in a system or network. & A machine or computer program. & 0.8154 \\
\hline
This can involve a lot of trial and error, which is why hackers might seem to be "jamming on their keyboards" as they try different approaches. & GLaDOS however , will give you no cake. & 0.7472 \\
\hline
A hacker is someone who uses their computer skills to try to gain access to systems or networks without permission. & Hackers are the people that get extra cake by going around the building and back through the door. & 0.8677 \\
\hline
This can involve a lot of trial and error, which is why hackers might seem to be "jamming on their keyboards" as they try different approaches. & I 've always wanted to know why hackers are jamming on their keyboards In reality , this does n't happen. & 0.8641 \\
\hline
They might also use tools to try to guess passwords or to find ways to get around security measures. & If you tell a computer to give a cake to every person that walks through the door , it will do. & 0.7735 \\
\hline
Hacking can involve a lot of typing and computer use, because hackers often use special software and programs to try to find weaknesses in a system or network. & Real computer hacking involves staring at a computer screen for hours of a time , searching a lot on Google , muttering " hmmm " and various expletives to oneself now and then , and stroking one '. & 0.8846 \\
\hline
Hackers might do this for a variety of reasons, such as to steal information, to cause damage or disruption, or just for the challenge of it. & They change the behavior of the subject to something that was never intended or even thought it would be possible by the creator of the subject. & 0.7937 \\
\hline
Hackers might do this for a variety of reasons, such as to steal information, to cause damage or disruption, or just for the challenge of it. & This is done in movies to make it look dramatic and exciting. & 0.7676 \\
\hline
\end{tabular}
\caption{This table depicts the sentence-level comparison of responses to Question 1 \hyperlink{Q1}{Q1}, given by a human \hyperlink{H1}{H1} and a Large Language Model \hyperlink{A1}{A1}. The cosine similarity values, derived from embeddings, represent the highest similarity between each pair of sentences in the human and LLM responses.}
\label{tab:appendix-table}
\end{table*}
This added transparency makes it easier for human evaluators to interpret the results and contributes to the elimination of the black-box nature often associated with existing AI text detection methods. To summarize, our method:
\begin{itemize}[noitemsep]
    \item Effectively generates diverse paraphrased questions using an advanced paraphrasing model.
    \item Produces accurate and contextually appropriate answers with the state-of-the-art LLM.
    \item Provides a comprehensive and transparent sentence-level evaluation, enabling the detection of subtle instances of plagiarism that might be overlooked by traditional methods.
\end{itemize}


\section{Conclusion}
In conclusion, this research presents a novel and effective method for detecting machine-generated text in academic settings, offering a valuable contribution to the field of plagiarism detection. By leveraging a comprehensive comparison technique, our approach provides more accurate and explainable evaluations compared to existing methods. The sentence level quantifiable metrics facilitate easier interpretation for human evaluators, mitigating the black-box nature of existing AI text detection methods.

Our model is adaptable to various NLG models, including cutting-edge LLMs like BardAI and Character.AI, ensuring its relevance and effectiveness as technology continues to evolve. This adaptability makes our approach a significant asset in maintaining academic integrity in the face of rapidly advancing natural language processing technologies.

Future research directions include collecting additional unbiased datasets for evaluation and comparing the performance of our model with other detection tools. We also plan to explore the incorporation of different algorithms at the sentence level, assembling them to achieve even better performance. Moreover, we plan to employ stylometry techniques to identify each student's unique writing style as more data from their responses are collected. This process will create a distinct signature based on the student's writing patterns, making it increasingly easy to detect plagiarism in future submissions.

These efforts will further refine our model and contribute to the ongoing pursuit of robust, transparent, and adaptable plagiarism detection methods in academia.

\bibliography{anthology,custom}
\bibliographystyle{acl_natbib}


\appendix

\end{document}